\documentclass{article}
\usepackage{spconf,amsmath,graphicx}
\usepackage{amssymb,amsfonts,booktabs,multirow}
\usepackage[hidelinks]{hyperref}

\title{Polarimetric Scattering-Structure Contrastive Learning for Few-Shot HRRP Recognition}

\name{Fangyi Wang, Peng Zhao, and Haipeng Wang%
\sthanks{Corresponding author: Haipeng Wang.}%
\sthanks{This work was supported in part by the National Natural Science Foundation of China (Grant No. 62271153) and the Fundamental and Interdisciplinary Disciplines Breakthrough Plan of the Ministry of Education of China (Grant No. JYB2025XDXM110).}%
\thanks{This work has been submitted to the IEEE for possible publication. Copyright may be transferred without notice, after which this version may no longer be accessible.}}
\address{Fudan University, China}

\begin{document}
\addtolength{\textheight}{0.15in}

\ninept
\maketitle

\begin{abstract}
Few-shot high-resolution range profile (HRRP) recognition is challenged by
strong aspect-dependent variation and sparse novel-class support. We propose
polarimetric scattering-structure contrastive learning (PSS-CL), which supplies
an episodic HRRP encoder with a structured physical training view.
Dominant local responses are represented by their soft affinities to canonical
polarimetric mechanisms, normalized range locations, and energy contributions.
A symmetric cross-view objective transfers this scattering structure to the
learned HRRP representation, while episodic prototype learning supports
novel-class inference without the physical branch. We conduct Random-aspect and
Cross-aspect experiments on a six-class measured dataset. In Cross-aspect 1-,
3-, and 5-shot recognition, PSS-CL achieves 81.73\%, 86.92\%, and 90.98\% mean
class accuracy (MCA), respectively, outperforming the best competing results by
5.62\%, 3.09\%, and 3.56\%.
\end{abstract}

\begin{keywords}
high-resolution range profile, few-shot learning, radar signal processing,
contrastive learning, radar target recognition
\end{keywords}

\section{Introduction}
\label{sec:introduction}

\begin{figure}[!t]
	\centering
	\includegraphics[width=0.9\columnwidth]{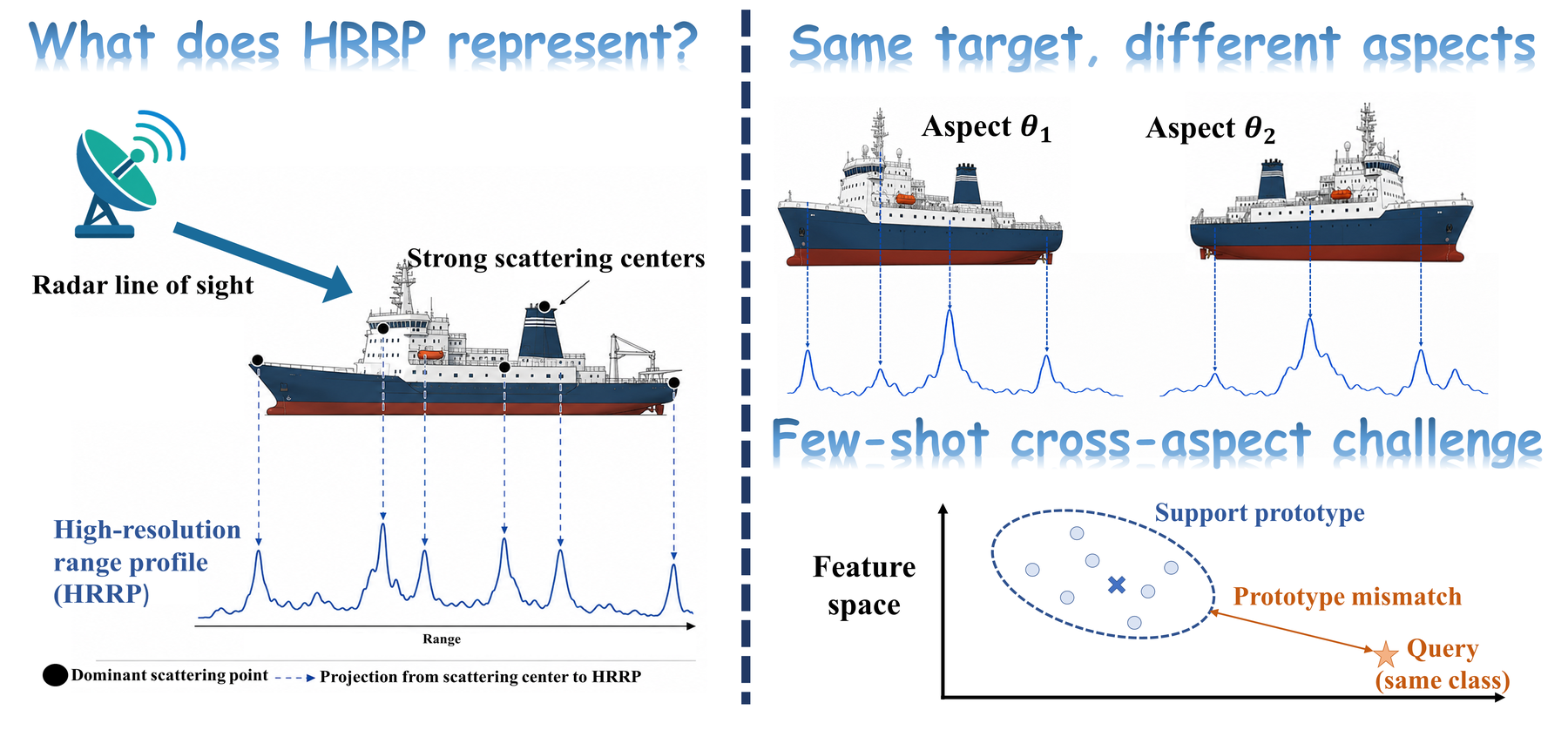}
	\caption{HRRP formation and the Cross-aspect few-shot recognition challenge.
		Dominant scattering centers form a one-dimensional range profile, while
		aspect-dependent scattering causes support--query prototype mismatch.}
\label{fig:intro_motivation}
\end{figure}

A high-resolution range profile (HRRP) is a one-dimensional complex radar
signature obtained by coherently projecting the scattering responses of a
target onto the radar line of sight. Its range cells resolve the distribution
of dominant scattering centers along the target extent, providing a compact
one-dimensional representation for recognition. Observation aspect changes
the visibility, amplitude, and phase of local scattering centers, resulting in
large intra-class variations. In few-shot recognition, one or a few labeled
support HRRPs provide sparse coverage of the aspect-dependent distribution of
each novel class. A query observed at a different aspect may consequently
deviate considerably from the support prototype. Figure~\ref{fig:intro_motivation}
summarizes HRRP formation and the resulting Cross-aspect prototype mismatch.

\begin{figure*}[t]
	\centering
	\includegraphics[width=0.90\textwidth]{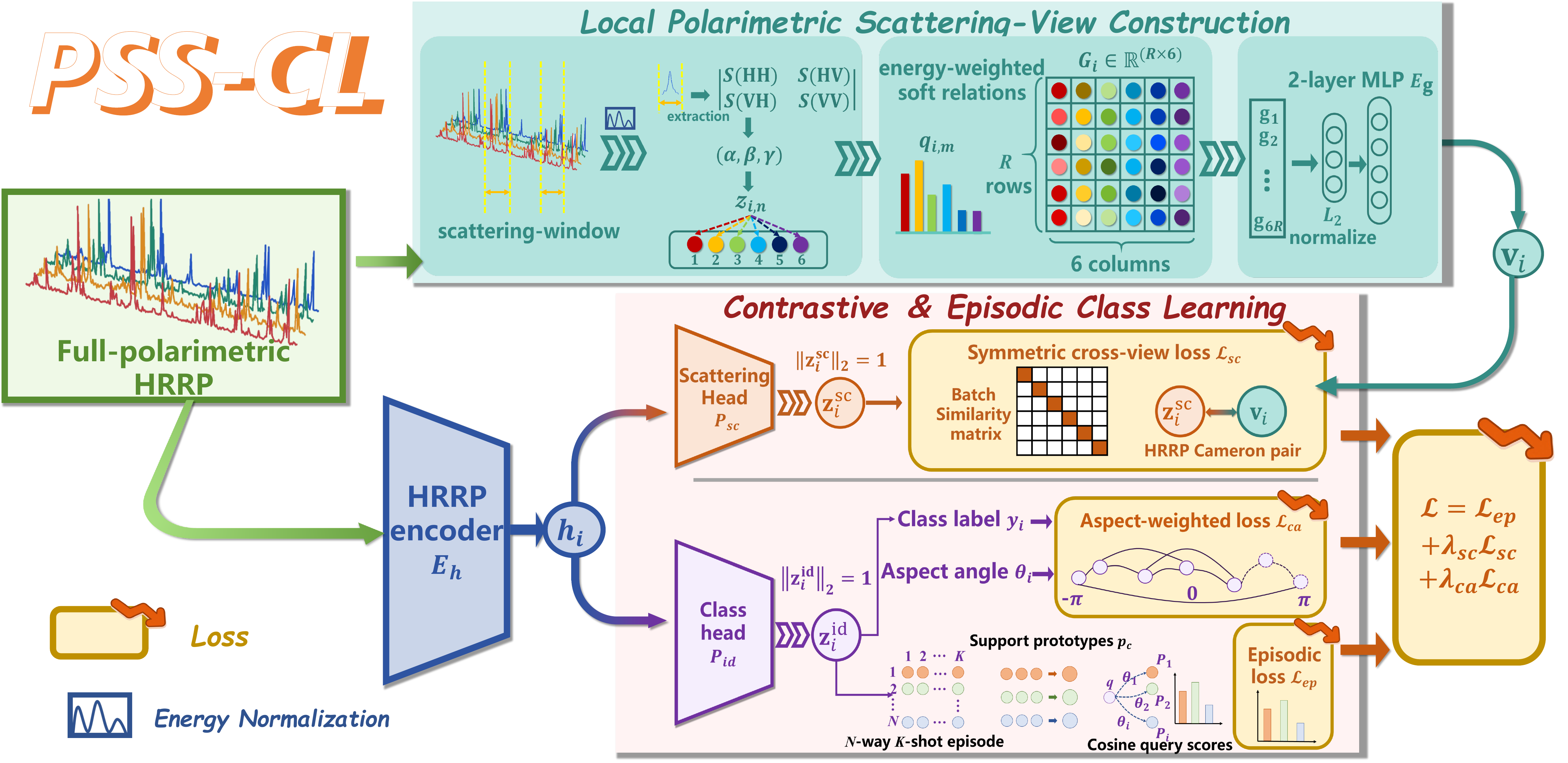}
	\caption{Overview of PSS-CL. The structured physical branch provides
	training-only cross-view supervision for episodic HRRP learning.}
	\label{fig:main}
\end{figure*}

Existing HRRP recognition studies have progressively replaced handcrafted
statistics with learned representations. Probabilistic and generative models
describe HRRPs using class-factorized, conditional, mixture, or adversarial
latent variables~\cite{Liao2021,Chen2020,Du2019,Huang2021}. Recurrent Bayesian
networks and deep sequence models exploit dependencies among multi-aspect
observations~\cite{Chen2021a,Guo2020}. More recent methods incorporate
prior-guided contrastive learning, semantic hierarchies, target-aspect
constraints, and polarization fusion to strengthen discriminative
representations~\cite{liu2024priorhrrp,liu2024sdhc,liu2025tagnet,
zhou2025dualpolarization}. These methods learn primarily from densely labeled
base-class distributions, leaving novel-class support sets with sparse aspect
coverage and unstable few-shot prototypes.

Few-shot learning transfers reusable knowledge from base classes to tasks with
small support sets. Representative
approaches learn similarity functions or task-specific metrics through Siamese
Networks, Matching Networks, Prototypical Networks, and Relation
Networks~\cite{koch2015siamese,vinyals2016matching,snell2017prototypical,
sung2018relation}; other methods learn adaptable initializations or episodic
classifiers~\cite{finn2017maml,lee2019metaoptnet}, and strong representation
baselines are also effective~\cite{chen2019closer}. For HRRP data,
prior-assisted task-wise discriminant analysis and fully polarimetric
scattering-center matching demonstrate the value of domain knowledge under
limited supervision~\cite{li2024taskwise,guo2025scmatching}. Existing objectives
and physical descriptors provide complementary cues, yet few-shot HRRP learning
lacks a fixed-dimensional training view that preserves local polarimetric
mechanism relations and range organization without altering novel-class
inference.

To address this problem, PSS-CL transfers a range-structured polarimetric
scattering view to an episodic HRRP encoder through cross-view contrast. An
auxiliary aspect-aware loss is used during training, while novel-class inference
requires only the HRRP encoder and prototype classifier.

Our main contributions are summarized as follows:
\begin{itemize}
\setlength{\topsep}{2pt}
\setlength{\itemsep}{1pt}
\setlength{\parsep}{0pt}
\setlength{\parskip}{0pt}
\item We construct a structured polarimetric scattering view by aggregating
canonical soft-response affinities, range locations, and energy contributions
from dominant scattering regions.
\item We introduce cross-view scattering alignment that transfers this physical
view to an episodically trained HRRP encoder. The physical branch supplies
training supervision and is removed for novel-class prototype inference.
\item On a six-class measured dataset, PSS-CL achieves the highest MCA among the
compared methods under the present protocol: 81.73\%, 86.92\%, and 90.98\% in
Cross-aspect 1-, 3-, and 5-shot recognition.
\end{itemize}
\section{Method}
\label{sec:method}
\subsection{Problem formulation and overview}

Let the base-class training set be
$\mathcal D_{\mathrm{base}}=\{(X_i,y_i,\theta_i)\}_{i=1}^{N_b}$,
where $X_i\in\mathbb R^{6\times L}$ is a fully polarimetric complex HRRP represented by the real and imaginary
parts of the $\mathrm{HH}$, $\mathrm{HV}$, and $\mathrm{VV}$ channels. The variables $y_i$ and
$\theta_i$ denote the class label and aspect angle, respectively. We consider
class-disjoint few-shot recognition: base classes are used for representation
learning; evaluation uses unseen classes under an $N$-way $K$-shot protocol.

PSS-CL combines structured scattering-view construction, cross-view alignment,
and episodic class learning. Cameron decomposition is applied to
dominant scattering regions; mechanism similarities, range locations, and
energy contributions form a compact physical view. A shared HRRP encoder
produces class and scattering embeddings. Cross-view contrast aligns the
scattering embedding with the physical view. Episodic prototypes optimize the
class embedding, with aspect-aware contrast serving as an auxiliary regularizer.
Inference uses only the encoder and class branch.

\subsection{Local polarimetric scattering-view construction}

For sample $i$, let $\mathbf S_i(n)$ be the $2\times2$ polarimetric scattering
matrix at range cell $n$. Its energy is
\begin{equation}
P_i(n)=\|\mathbf S_i(n)\|_F^2
=|S_{i,\mathrm{HH}}(n)|^2+2|S_{i,\mathrm{HV}}(n)|^2
+|S_{i,\mathrm{VV}}(n)|^2,
\label{eq:pol_energy}
\end{equation}
where calibrated monostatic reciprocity gives
$S_{i,\mathrm{VH}}=S_{i,\mathrm{HV}}$. After joint energy normalization, a
seven-cell moving average and $0.1$-normalized prominence threshold with
seven-cell non-maximum suppression retain $M_i\leq M_{\max}=16$ seven-cell
windows; the global maximum handles an empty detection. Their centers and
energies are
\begin{equation}
c_{i,m}=\arg\max_{n\in\mathcal W_{i,m}}P_i(n),\qquad
A_{i,m}=\sum_{n\in\mathcal W_{i,m}}P_i(n).
\label{eq:window_stat}
\end{equation}

\begin{figure}[t]
\centering
\includegraphics[width=0.7\columnwidth,trim=0 90 0 0,clip]
{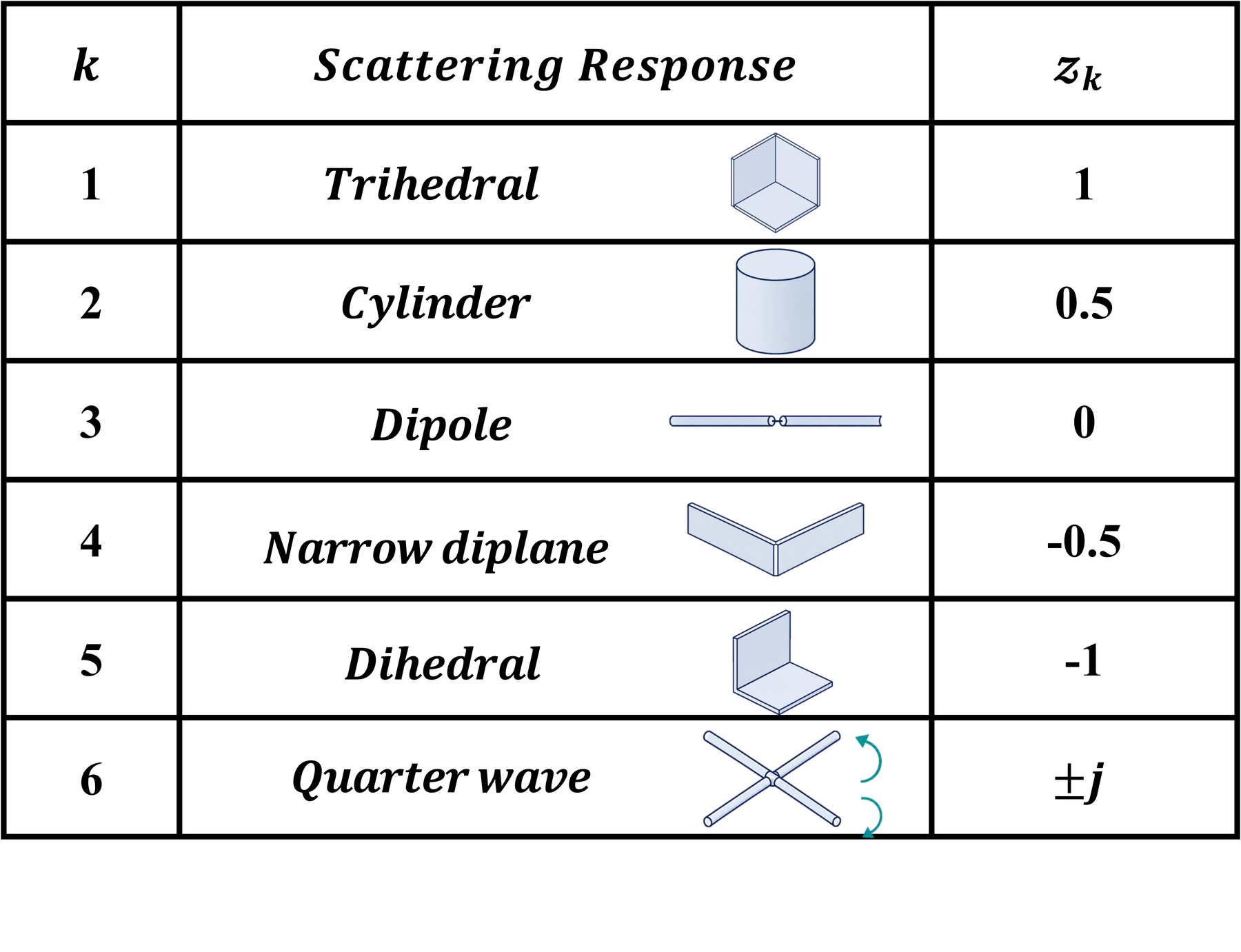}
\caption{Six canonical polarimetric scattering-response families used in the
Cameron implementation.}
\label{fig:scattering_responses}
\end{figure}

We instantiate the local polarimetric decomposition using the Cameron
formulation, which characterizes a reciprocal response by its maximum symmetric
component and measures it against canonical scattering
responses~\cite{Cameron1990,Cameron1996,CameronRais2009}. For the scattering
matrix at cell $n$, define
$\alpha=(S_{\mathrm{HH}}+S_{\mathrm{VV}})/\sqrt{2}$,
$\beta=(S_{\mathrm{HH}}-S_{\mathrm{VV}})/\sqrt{2}$, and
$\gamma=(S_{\mathrm{HV}}+S_{\mathrm{VH}})/\sqrt{2}$, where sample and cell
indices are omitted. The Cameron real rotation parameter and maximum symmetric
component are
\begin{equation}
\begin{aligned}
\varphi&=\frac{1}{2}\operatorname{atan2}\!\left(
2\Re\{\beta\gamma^{*}\},|\beta|^2-|\gamma|^2\right),\\
\mu&=\beta\cos\varphi+\gamma\sin\varphi.
\end{aligned}
\label{eq:cameron_coordinate}
\end{equation}
Let $\lambda^{(1)}$ and $\lambda^{(2)}$ denote
$\alpha+\mu$ and $\alpha-\mu$, ordered such that
$|\lambda^{(1)}|\geq|\lambda^{(2)}|$. Restoring the indices, the Cameron
coordinate is $z_{i,n}=\lambda_{i,n}^{(2)}/\lambda_{i,n}^{(1)}$, with
$|z_{i,n}|\leq1$; it is set to zero when
$|\lambda_{i,n}^{(1)}|<\delta$, where $\delta=10^{-8}$. Figure~\ref{fig:scattering_responses} illustrates the six
canonical response families used to construct the local scattering view. Their
coordinates are $z_k\in\{1,0.5,0,-0.5,-1,\mathrm{j}\}$. The pair
$\pm\mathrm{j}$ constitutes one quarter-wave family under the Cameron metric.

The squared Cameron affinity, $s_{i,n,k}=\cos^2 d_C(z_{i,n},z_k)$, is
\begin{equation}
s_{i,n,k}=
\frac{\max\!\left(
|1+z_{i,n}z_k^*|^2,|z_{i,n}+z_k^*|^2\right)}
{(1+|z_{i,n}|^2)(1+|z_k|^2)},
\quad k=1,\ldots,6,
\label{eq:cameron_similarity}
\end{equation}
where $s_{i,n,k}\in[0,1]$. Each retained cell is represented by its complete
six-dimensional similarity vector. The energy-weighted relation within window
$m$ is
\begin{equation}
q_{i,m,k}=
\frac{\sum_{n\in\mathcal W_{i,m}}P_i(n)s_{i,n,k}}
{\sum_{n\in\mathcal W_{i,m}}P_i(n)+\delta}.
\label{eq:cameron_soft}
\end{equation}
The vector $\mathbf q_{i,m}=[q_{i,m,1},\ldots,q_{i,m,6}]$ records the
continuous relation between a local response and all canonical mechanisms.

To obtain a fixed-dimensional representation, the span from the first to last
retained window boundary is mapped linearly to $[0,1]$ and divided into $R$
intervals. Let $b(c_{i,m})\in\{1,\ldots,R\}$ denote the clipped interval index
of center $c_{i,m}$ and define
\begin{equation}
\begin{aligned}
w_{i,m}&=\frac{A_{i,m}}{\sum_{j=1}^{M_i}A_{i,j}+\delta},\\
G_i(r,k)&=\sum_{\substack{m=1\\b(c_{i,m})=r}}^{M_i}
w_{i,m}q_{i,m,k}.
\end{aligned}
\label{eq:structured_matrix}
\end{equation}
Collecting these entries gives
$\mathbf G_i=[G_i(r,k)]\in\mathbb R^{R\times6}$.
The structured and location-free global scattering views are
\begin{equation}
\mathbf g_i=\operatorname{norm}(\operatorname{vec}(\mathbf G_i))
\in\mathbb R^{6R},\quad
\bar{\mathbf g}_i=\operatorname{norm}\!\left(\sum_{m=1}^{M_i}
w_{i,m}\mathbf q_{i,m}\right).
\label{eq:scattering_view}
\end{equation}
Here, $\operatorname{norm}(\mathbf x)=\mathbf x/
\max(\|\mathbf x\|_2,\delta)$. Thus, $\mathbf g_i$ describes how polarimetric
mechanisms are distributed along range; $\bar{\mathbf g}_i$ removes their
locations for the ablation in Table~\ref{tab:ablation}.

\subsection{Cross-view scattering alignment and class learning}

A lightweight one-dimensional encoder extracts the shared HRRP feature
$\mathbf h_i=E_h(X_i)$. Two projection heads generate the class and scattering
embeddings, where ``id'' denotes the target-identity (class) space. A two-layer
multilayer perceptron encodes the structured scattering view:
\begin{equation}
\begin{split}
\mathbf z_i^{\mathrm{id}}&=\operatorname{norm}
\left(P_{\mathrm{id}}(\mathbf h_i)\right),\\
\mathbf z_i^{\mathrm{sc}}&=\operatorname{norm}
\left(P_{\mathrm{sc}}(\mathbf h_i)\right),\qquad
\mathbf v_i=\operatorname{norm}\left(E_g(\mathbf g_i)\right).
\end{split}
\label{eq:embeddings}
\end{equation}
The paired embeddings $\mathbf z_i^{\mathrm{sc}}$ and $\mathbf v_i$ describe
the same HRRP sample in a training episode batch of size $B$. With
$\ell_{ij}^{\mathrm{sc}}=(\mathbf z_i^{\mathrm{sc}})^\top\mathbf v_j/
\tau_{\mathrm{sc}}$, their symmetric cross-view contrastive loss is
\begin{equation}
\mathcal L_{\mathrm{sc}}=-\frac{1}{2B}\sum_{i=1}^{B}
\left[
\log\frac{e^{\ell_{ii}^{\mathrm{sc}}}}
{\sum_{j=1}^{B}e^{\ell_{ij}^{\mathrm{sc}}}}
+
\log\frac{e^{\ell_{ii}^{\mathrm{sc}}}}
{\sum_{j=1}^{B}e^{\ell_{ji}^{\mathrm{sc}}}}
\right].
\label{eq:scattering_contrast}
\end{equation}
This objective transfers local polarimetric structure to the shared HRRP
feature. Same-class instances remain negatives because $\mathbf g_i$ is
sample-specific; class invariance is learned only by the identity head.

As an auxiliary class-space regularizer during base-class training, the class
branch employs aspect-aware supervised contrastive learning~\cite{liu2024priorhrrp}.
For anchor $i$, let
$\mathcal P(i)=\{p\ne i\mid y_p=y_i\}$. Each contrastive batch contains all
support and query samples in an episode, ensuring $\mathcal P(i)\ne\varnothing$.
The circular aspect difference and normalized positive weight are
\begin{equation}
\begin{split}
\Delta\theta_{ip}&=\min\left(|\theta_i-\theta_p|,
360^{\circ}-|\theta_i-\theta_p|\right),\\
\widetilde a_{ip}&=
\frac{1+\eta\Delta\theta_{ip}/180^{\circ}}
{\sum_{r\in\mathcal P(i)}
(1+\eta\Delta\theta_{ir}/180^{\circ})}.
\end{split}
\label{eq:aspect_weight}
\end{equation}
The coefficient $\eta\geq0$ controls the emphasis on wider aspect separations.
The resulting loss is
\begin{equation}
\mathcal L_{\mathrm{ca}}=-\frac{1}{B}\sum_{i=1}^{B}
\sum_{p\in\mathcal P(i)}\widetilde a_{ip}
\log\frac{
e^{(\mathbf z_i^{\mathrm{id}})^\top\mathbf z_p^{\mathrm{id}}/
\tau_{\mathrm{ca}}}}
{\sum_{j\ne i}
e^{(\mathbf z_i^{\mathrm{id}})^\top\mathbf z_j^{\mathrm{id}}/
\tau_{\mathrm{ca}}}}.
\label{eq:aspect_contrast}
\end{equation}
The weighting emphasizes same-class positive pairs separated by wider aspect
intervals. Aspect angles determine positive-pair weights only during training;
the HRRP signal $X_i$ remains the input to $E_h$.

\subsection{Episodic prototype learning and inference}

For an $N$-way $K$-shot episode with query set $\mathcal Q$, the prototype,
score, episodic loss, and total objective are
\begin{equation}
\begin{gathered}
\mathbf p_c=\operatorname{norm}\!\left(K^{-1}\sum_{k=1}^{K}
\mathbf z_{c,k}^{\mathrm{id}}\right),\qquad
d(q,c)=(\mathbf z_q^{\mathrm{id}})^\top\mathbf p_c,\\
\mathcal L_{\mathrm{ep}}=-|\mathcal Q|^{-1}\!\sum_{(q,y_q)\in\mathcal Q}
\log\frac{e^{d(q,y_q)/\tau_{\mathrm{ep}}}}
{\sum_c e^{d(q,c)/\tau_{\mathrm{ep}}}},\\
\mathcal L=\mathcal L_{\mathrm{ep}}+
\lambda_{\mathrm{sc}}\mathcal L_{\mathrm{sc}}+
\lambda_{\mathrm{ca}}\mathcal L_{\mathrm{ca}}.
\end{gathered}
\label{eq:learning_objectives}
\end{equation}
Novel-class evaluation encodes support and query HRRPs with $E_h$ and
$P_{\mathrm{id}}$, constructs support prototypes, and applies cosine scoring.
The scattering-view encoder, scattering head, and aspect annotations provide
training supervision and are discarded after training.

\section{Experiments}
\label{sec:Experiment}

\subsection{Dataset and settings}

The measured dataset contains 56,356 complex HRRPs in three synchronized
channels: $\mathrm{HH}$, $\mathrm{HV}$, and $\mathrm{VV}$. The range-registered
profiles come from six targets: bulk carrier, container, passenger, fishing,
tanker, and reflector group. Each
$L=1024$ HRRP was acquired at 10~GHz with a 120-MHz bandwidth and 1.25-m range
resolution. For each class, the measurements collectively span
$0^{\circ}$--$360^{\circ}$ in azimuth and $0^{\circ}$--$25^{\circ}$ in pitch.
We use two complementary 3/3 class splits: Fold A assigns the first three
classes to base training and the remaining classes to novel-class evaluation;
Fold B exchanges the two groups.

Each fold trains one full-azimuth base-class model for both protocols. A 3-way
$K$-shot episode, $K\in\{1,3,5\}$, contains 15 queries per class, with no overlap
between support and query samples. Random-aspect draws samples without
replacement from the full azimuth range. Cross-aspect draws support and query
samples from two $60^{\circ}$ sectors centered $120^{\circ}$ apart, leaving a
$60^{\circ}$ boundary gap. Each fold uses 5,000 training episodes with uniformly
sampled $K$ and 500 episodes from a disjoint base-class validation pool. All
methods share 2,000 test episodes per fold, protocol, and shot. MCA first
averages per-class query accuracy over episodes and then over folds.

The four-stage 1-D ResNet-12 outputs 128-dimensional embeddings after global
average pooling; all three heads are two-layer MLPs. One episode forms a batch.
We set $R=16$,
$\tau_{\mathrm{sc}}=\tau_{\mathrm{ca}}=0.07$, $\tau_{\mathrm{ep}}=0.1$,
$\eta=1$, and $\lambda_{\mathrm{sc}}=\lambda_{\mathrm{ca}}=0.5$. AdamW uses a
$10^{-3}$ learning rate and $10^{-4}$ weight decay; all choices use only base
validation.

\begin{table}[!h]
	\caption{MCA (\%) on the measured HRRP dataset.}
	\label{tab:main_results}
	\centering
	\setlength{\tabcolsep}{2.5pt}
	\renewcommand{\arraystretch}{0.90}
	
	\begin{tabular*}{\columnwidth}{@{\extracolsep{\fill}}lcccccc@{}}
		\toprule
		\multirow{2}{*}{Method}
		& \multicolumn{3}{c}{Random-aspect}
		& \multicolumn{3}{c}{Cross-aspect} \\
		\cmidrule(lr){2-4}\cmidrule(lr){5-7}
		& 1-shot & 3-shot & 5-shot
		& 1-shot & 3-shot & 5-shot \\
		\midrule
		
		\multicolumn{7}{@{}l}{\itshape General few-shot learning} \\
		Matching Nets~\cite{vinyals2016matching} & 72.42 & 79.16 & 84.28 & 60.83 & 67.76 & 70.69 \\
		ProtoNet~\cite{snell2017prototypical} & 76.28 & 82.36 & 84.55 & 63.52 & 72.32 & 73.85 \\
		RelationNet~\cite{sung2018relation} & 74.76 & 81.55 & 85.39 & 63.19 & 70.93 & 75.46 \\
		MAML~\cite{finn2017maml} & 70.35 & 74.16 & 76.60 & 58.70 & 63.98 & 67.25 \\
		MetaOptNet~\cite{lee2019metaoptnet} & 75.95 & 83.05 & 86.16 & 66.81 & 74.72 & 76.66 \\
		DN4~\cite{li2019dn4} & 77.53 & 82.73 & 87.13 & 67.47 & 75.69 & 77.53 \\
		FEAT~\cite{ye2020feat} & 76.72 & 81.55 & 83.64 & 68.87 & 74.92 & 80.52 \\
		DeepEMD~\cite{zhang2020deepemd} & 81.90 & 83.89 & 86.56 & 71.20 & 75.51 & 80.10 \\
		
		\midrule
		\multicolumn{7}{@{}l}{\itshape HRRP-oriented methods} \\
		PriorK-NN~\cite{liu2024priorhrrp} & 73.61 & 79.73 & 85.49 & 69.42 & 76.10 & 79.23 \\
		GAF-MLGNN~\cite{chen2026gafmlgnn} & 80.88 & 85.71 & 90.81 & 73.93 & 80.65 & 84.71 \\
		MTBC~\cite{li2025mtbc} & 85.71 & 89.27 & 92.65 & 76.11 & 82.39 & 87.42 \\
		DeepSCOM~\cite{guo2025scmatching} & 82.05 & 86.40 & 88.94 & 73.85 & 83.83 & 86.99 \\
		
		\midrule
		PSS-CL (Ours)
		& \textbf{88.23}
		& \textbf{93.09}
		& \textbf{95.21}
		& \textbf{81.73}
		& \textbf{86.92}
		& \textbf{90.98} \\
		\bottomrule
	\end{tabular*}
\end{table}

\subsection{Comparison with competing methods}

Table~\ref{tab:main_results} compares eight general and four HRRP-oriented
few-shot methods. All use the same folds, episodes, inputs, and support--query
selections. General baselines share ResNet-12, while HRRP methods retain their
published architectures. PSS-CL ranks first in every
setting, exceeding the best competing result by 2.52\%, 3.82\%, and 2.56\% under
Random-aspect evaluation and by 5.62\%, 3.09\%, and 3.56\% under Cross-aspect
evaluation. Its 1-shot Random-to-Cross decrease is 6.50\%, compared with 12.76\%
for ProtoNet. The smaller drop suggests that scattering-structure supervision
helps preserve class information when the support and query aspects differ.

\begin{figure}[!h]
\centering
\includegraphics[width=\columnwidth]
{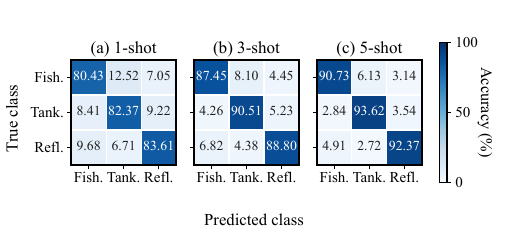}
\caption{Cross-aspect confusion matrices for the fishing--tanker--reflector fold:
(a) 1-shot; (b) 3-shot; (c) 5-shot.}
\label{fig:confusion}
\end{figure}

Figure~\ref{fig:confusion} reports the Cross-aspect fishing--tanker--reflector
fold. The mean diagonal accuracy rises from 82.14\% to 88.92\% and 92.24\% as
the shot number increases. At 5-shot, all three class accuracies exceed 90\%,
compared with 80.43\%--83.61\% at 1-shot. Fishing-to-tanker errors fall from
12.52\% to 8.10\% and 6.13\%, while ship--reflector errors fall from
6.71\%--9.68\% to 2.72\%--4.91\%. The consistent decrease shows that additional
support samples improve the prototypes for all three classes.

\subsection{Ablation study}

Table~\ref{tab:ablation} separates the effects of cross-view alignment, aspect
regularization, and the structured view $\mathbf g_i$. With the global view
$\bar{\mathbf g}_i$, $\mathcal L_{\mathrm{sc}}$ improves the episodic baseline
by 10.64\%, 9.36\%, and 8.69\%, while $\mathcal L_{\mathrm{ca}}$ alone adds
7.26\%, 6.10\%, and 5.79\%. Combining both losses reaches 78.33\%, 84.49\%, and
87.68\%. Replacing $\bar{\mathbf g}_i$ with the range-structured view
$\mathbf g_i$ further adds 3.40\%, 2.43\%, and 3.30\%. Overall, PSS-CL improves
the baseline by 17.23\%, 14.53\%, and 14.16\%. The larger 1-shot gain highlights
the value of physical structure when support is most limited, while the
consistent gains show that the three components work together.

\begin{table}[!h]
\caption{Ablation results under Cross-aspect evaluation (MCA, \%).}
\label{tab:ablation}
\centering
\setlength{\tabcolsep}{2.6pt}
\renewcommand{\arraystretch}{1.06}
\begin{tabular}{lcccccc}
\toprule
Configuration & $\mathcal L_{\mathrm{sc}}$ & $\mathcal L_{\mathrm{ca}}$
& View & 1-shot & 3-shot & 5-shot \\
\midrule
Episodic baseline & -- & -- & -- & 64.50 & 72.39 & 76.82 \\
Scattering alignment & \checkmark & -- & $\bar{\mathbf g}_i$ & 75.14 & 81.75 & 85.51 \\
Aspect regularizer & -- & \checkmark & -- & 71.76 & 78.49 & 82.61 \\
Global scattering view & \checkmark & \checkmark & $\bar{\mathbf g}_i$ & 78.33 & 84.49 & 87.68 \\
PSS-CL (Ours) & \checkmark & \checkmark & $\mathbf g_i$
& \textbf{81.73} & \textbf{86.92} & \textbf{90.98} \\
\bottomrule
\end{tabular}
\end{table}

\subsection{Hyperparameter sensitivity}

\begin{figure}[!h]
\centering
\includegraphics[width=0.95\columnwidth]
{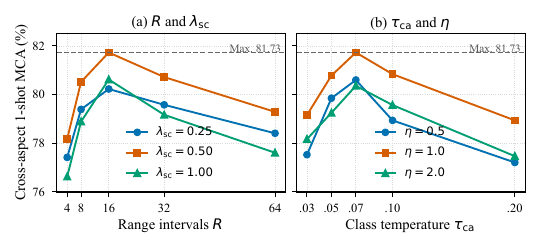}
\caption{Post-selection Cross-aspect 1-shot sensitivity to (a) $R$ and
$\lambda_{\mathrm{sc}}$; (b) $\tau_{\mathrm{ca}}$ and $\eta$.}
\label{fig:sensitivity}
\end{figure}

Figure~\ref{fig:sensitivity} reports a post-selection Cross-aspect 1-shot
diagnostic, with defaults fixed on the base validation set. For
$\lambda_{\mathrm{sc}}=0.5$, the MCA rises from 78.18\% at $R=4$ to 81.73\% at
$R=16$, then falls to 79.29\% at $R=64$. This agrees with the need to balance
range detail against sparsity. At $R=16$, the three tested values of
$\lambda_{\mathrm{sc}}$ give 80.23\%--81.73\%. For the aspect regularizer,
$\tau_{\mathrm{ca}}=0.07$ gives 80.60\%, 81.73\%, and 80.36\% for
$\eta=0.5$, 1, and 2, whereas all curves fall below 79\% at
$\tau_{\mathrm{ca}}=0.20$. The gradual changes around the selected values show
that performance does not rely on a narrow parameter choice.

\section{Conclusion}
\label{sec:conclusion}

In this paper, we presented PSS-CL for few-shot HRRP recognition. Cross-view
contrast transfers range--mechanism organization from a structured fully
polarimetric view to the HRRP encoder. Episodic prototype learning and aspect
regularization support novel-class recognition, with the physical branch used
only during training. On the measured six-class dataset, PSS-CL achieves the
highest MCA among the compared methods under both protocols.

\clearpage

\bibliographystyle{IEEEbib}
\bibliography{strings,icassp_hrrp}

@InProceedings{Cameron1990,
  author    = {W. L. Cameron and L. K. Leung},
  title     = {Feature Motivated Polarization Scattering Matrix Decomposition},
  booktitle = {Proc. IEEE Int. Radar Conf.},
  year      = {1990},
  pages     = {549--557},
  doi       = {10.1109/RADAR.1990.201088},
}

@Article{Cameron1996,
  author  = {W. L. Cameron and N. N. Youssef and L. K. Leung},
  title   = {Simulated Polarimetric Signatures of Primitive Geometrical Shapes},
  journal = {IEEE Trans. Geosci. Remote Sens.},
  year    = {1996},
  volume  = {34},
  number  = {3},
  pages   = {793--803},
  doi     = {10.1109/36.499784},
}

@Article{CameronRais2009,
  author  = {W. L. Cameron and H. Rais},
  title   = {Polarization Symmetric Scatterer Metric Space},
  journal = {IEEE Trans. Geosci. Remote Sens.},
  year    = {2009},
  volume  = {47},
  number  = {4},
  pages   = {1097--1107},
  doi     = {10.1109/TGRS.2008.2007962},
}

@InProceedings{koch2015siamese,
  author    = {G. Koch and R. Zemel and R. Salakhutdinov},
  booktitle = {ICML Deep Learning Workshop},
  title     = {Siamese Neural Networks for One-Shot Image Recognition},
  year      = {2015},
}

@InProceedings{vinyals2016matching,
  author    = {O. Vinyals and C. Blundell and T. Lillicrap and K. Kavukcuoglu and D. Wierstra},
  booktitle = {Adv. Neural Inf. Process. Syst.},
  title     = {Matching Networks for One Shot Learning},
  year      = {2016},
  volume    = {29},
}

@InProceedings{finn2017maml,
  author    = {C. Finn and P. Abbeel and S. Levine},
  booktitle = {Proc. 34th Int. Conf. Mach. Learn. (ICML)},
  title     = {Model-Agnostic Meta-Learning for Fast Adaptation of Deep Networks},
  year      = {2017},
  pages     = {1126--1135},
  volume    = {70},
}

@InProceedings{snell2017prototypical,
  author    = {J. Snell and K. Swersky and R. Zemel},
  booktitle = {Adv. Neural Inf. Process. Syst.},
  title     = {Prototypical Networks for Few-Shot Learning},
  year      = {2017},
  volume    = {30},
}

@InProceedings{sung2018relation,
  author    = {F. Sung and Y. Yang and L. Zhang and T. Xiang and P. H. S. Torr and T. M. Hospedales},
  booktitle = {Proc. IEEE Conf. Comput. Vis. Pattern Recognit. (CVPR)},
  title     = {Learning to Compare: Relation Network for Few-Shot Learning},
  year      = {2018},
  pages     = {1199--1208},
}

@InProceedings{chen2019closer,
  author    = {W. Y. Chen and Y. C. Liu and Z. Kira and Y. C. F. Wang and J. B. Huang},
  booktitle = {Int. Conf. Learn. Represent. (ICLR)},
  title     = {A Closer Look at Few-Shot Classification},
  year      = {2019},
}

@InProceedings{lee2019metaoptnet,
  author    = {K. Lee and S. Maji and A. Ravichandran and S. Soatto},
  booktitle = {Proc. IEEE/CVF Conf. Comput. Vis. Pattern Recognit. (CVPR)},
  title     = {Meta-Learning with Differentiable Convex Optimization},
  year      = {2019},
  pages     = {10657--10665},
}

@InProceedings{li2019dn4,
  author    = {W. Li and L. Wang and J. Xu and J. Huo and Y. Gao and J. Luo},
  booktitle = {Proc. IEEE/CVF Conf. Comput. Vis. Pattern Recognit. (CVPR)},
  title     = {Revisiting Local Descriptor Based Image-to-Class Measure for Few-Shot Learning},
  year      = {2019},
  pages     = {7260--7268},
}

@InProceedings{ye2020feat,
  author    = {H. J. Ye and H. Hu and D. C. Zhan and F. Sha},
  booktitle = {Proc. IEEE/CVF Conf. Comput. Vis. Pattern Recognit. (CVPR)},
  title     = {Few-Shot Learning via Embedding Adaptation with Set-to-Set Functions},
  year      = {2020},
  pages     = {8808--8817},
}

@InProceedings{zhang2020deepemd,
  author    = {C. Zhang and Y. Cai and G. Lin and C. Shen},
  booktitle = {Proc. IEEE/CVF Conf. Comput. Vis. Pattern Recognit. (CVPR)},
  title     = {{DeepEMD}: Few-Shot Image Classification with Differentiable Earth Mover's Distance and Structured Classifiers},
  year      = {2020},
  pages     = {12203--12213},
}

@Article{liu2024priorhrrp,
  author  = {Q. Liu and X. Zhang and Y. Liu},
  journal = {IEEE Trans. Aerosp. Electron. Syst.},
  title   = {A Prior-Knowledge-Guided Neural Network Based on Supervised Contrastive Learning for Radar {HRRP} Recognition},
  year    = {2024},
  volume  = {60},
  number  = {3},
  pages   = {2854--2873},
  doi     = {10.1109/TAES.2024.3353718},
}

@Article{liu2024sdhc,
  author  = {Y. Liu and T. Long and L. Zhang and Y. Wang and X. Zhang and Y. Li},
  journal = {IEEE Trans. Aerosp. Electron. Syst.},
  title   = {{SDHC}: Joint Semantic-Data Guided Hierarchical Classification for Fine-Grained {HRRP} Target Recognition},
  year    = {2024},
  volume  = {60},
  number  = {4},
  pages   = {3993--4009},
  doi     = {10.1109/TAES.2024.3373378},
}

@Article{li2024taskwise,
  author  = {J. Li and W. Guo and F. Wei and T. Zhang and W. Yu},
  journal = {IEEE Trans. Aerosp. Electron. Syst.},
  title   = {Prior Information-Assisted Few-Shot {HRRP} Recognition Based on Task-Wise Shrinkage Quadratic Discriminant Analysis},
  year    = {2024},
  volume  = {60},
  number  = {6},
  pages   = {9354--9368},
  doi     = {10.1109/TAES.2024.3441548},
}

@Article{guo2025scmatching,
  author  = {Z. Guo and Z. Liu and R. Xie and L. Ran},
  journal = {IEEE Trans. Aerosp. Electron. Syst.},
  title   = {{HRRP} Few-Shot Target Recognition for Full Polarimetric Radars via {SCs} Optimal Matching},
  year    = {2025},
  volume  = {61},
  number  = {2},
  pages   = {4526--4541},
  doi     = {10.1109/TAES.2024.3505117},
}

@Article{zhou2025dualpolarization,
  author  = {Y. Zhou and S. Liu and H. W. Gao and H. Lin and G. Wei and X. Wang and X. M. Pan},
  journal = {IEEE Trans. Aerosp. Electron. Syst.},
  title   = {A Dual-Polarization Feature Fusion Network for Radar Automatic Target Recognition Based on {HRRP} Sequences},
  year    = {2025},
  volume  = {61},
  number  = {5},
  pages   = {14642--14655},
  doi     = {10.1109/TAES.2025.3586217},
}

@Article{liu2025tagnet,
  author  = {Q. Liu and X. Zhang and Y. Liu},
  journal = {IEEE Trans. Aerosp. Electron. Syst.},
  title   = {Target-Aspect-Guided Neural Network With Geometric Constraints for Imbalanced Radar Target Recognition},
  year    = {2025},
  volume  = {61},
  number  = {5},
  pages   = {12122--12142},
  doi     = {10.1109/TAES.2025.3573273},
}

@Article{li2025mtbc,
  author  = {S. Li and W. Li and P. Huang and M. Zheng and B. Tian and S. Xu},
  journal = {IEEE Sensors J.},
  title   = {{MTBC}: Masked Vision Transformer and Brownian Distance Covariance Classifier for Cross-Domain Few-Shot {HRRP} Recognition},
  year    = {2025},
  volume  = {25},
  number  = {9},
  pages   = {16440--16454},
  doi     = {10.1109/JSEN.2025.3550584},
}

@Article{chen2026gafmlgnn,
  author  = {L. Chen and P. Hu and Q. Liu and Z. Liu},
  journal = {IEEE Trans. Signal Inf. Process. Netw.},
  title   = {{GAF-MLGNN}: An Efficient Meta-Learning Framework for Few-Shot {HRRP} {RATR} With {GNN}},
  year    = {2026},
  volume  = {12},
  pages   = {341--356},
  doi     = {10.1109/TSIPN.2026.3663887},
}

@Article{Liao2021,
  author    = {L. Liao and L. Du and J. Chen},
  journal   = {Signal Process.},
  title     = {Class factorized complex variational auto-encoder for {HRR} radar target recognition},
  year      = {2021},
  pages     = {107932},
  volume    = {182},
  publisher = {Elsevier},
}

@Article{Huang2021,
  author    = {Y. Huang and Y. Wen and L. Shi and X. Ding},
  journal   = {IEEE Geosci. Remote Sens. Lett.},
  title     = {Recognition-aware {HRRP} generation with generative adversarial network},
  year      = {2022},
  pages     = {1--5},
  volume    = {19},
  publisher = {IEEE},
}

@Article{Chen2020,
  author    = {J. Chen and L. Du and L. Liao},
  journal   = {IEEE Trans. Cybern.},
  title     = {Discriminative mixture variational autoencoder for semisupervised classification},
  year      = {2022},
  number    = {5},
  pages     = {3032--3046},
  volume    = {52},
  publisher = {IEEE},
}

@Article{Du2019,
  author    = {C. Du and B. Chen and B. Xu and D. Guo and H. Liu},
  journal   = {Signal Process.},
  title     = {Factorized discriminative conditional variational auto-encoder for radar {HRRP} target recognition},
  year      = {2019},
  pages     = {176--189},
  volume    = {158},
  publisher = {Elsevier},
}

@Article{Chen2021a,
  author    = {W. Chen and B. Chen and Y. Liu and X. Peng and H. Fan and F. Yu and H. Liu},
  journal   = {Signal Process.},
  title     = {Bidirectional recurrent gamma belief network for {HRRP} target recognition},
  year      = {2021},
  pages     = {108213},
  volume    = {188},
  publisher = {Elsevier},
}

@Article{Guo2020,
  author    = {D. Guo and B. Chen and W. Chen and C. Wang and H. Liu and M. Zhou},
  journal   = {IEEE Trans. Signal Process.},
  title     = {Variational temporal deep generative model for radar {HRRP} target recognition},
  year      = {2020},
  pages     = {5795--5809},
  volume    = {68},
  publisher = {IEEE},
}

\begin{center}
{\bfseries DISCLOSURE}
\end{center}
\vspace{-.5em}
The authors have no relevant financial or nonfinancial interests to disclose.

\begin{center}
{\bfseries COMPLIANCE WITH ETHICAL STANDARDS}
\end{center}
\vspace{-.5em}
This study did not involve human or animal subjects; therefore,\\
ethical approval was not required.

\end{document}